\documentclass[sigconf]{acmart}

\usepackage{pifont}
\usepackage{colortbl}
\usepackage{multirow}
\usepackage{xspace}

\newcommand{\eg}{e.g.,\xspace}
\newcommand{\ie}{i.e.,\xspace}

\makeatletter
\DeclareRobustCommand\onedot{\futurelet\@let@token\@onedot}
\def\@onedot{\ifx\@let@token.\else.\null\fi\xspace}

\makeatother

\AtBeginDocument{%
  }

\copyrightyear{2026}
\acmYear{2026}
\setcopyright{cc}
\setcctype{by}
\acmConference[MM '26]{Proceedings of the 34th ACM International Conference on Multimedia}{November 10--14, 2026}{Rio de Janeiro, Brazil}
\acmBooktitle{Proceedings of the 34th ACM International Conference on Multimedia (MM '26), November 10--14, 2026, Rio de Janeiro, Brazil}
\acmDOI{10.1145/3767308.3835292}
\acmISBN{979-8-4007-2213-4/2026/11}

\acmSubmissionID{2263}

\begin{document}

\title{Layer Roles Matter: Mitigating Modality Imbalance in Multimodal Emotion Reasoning via Adaptive Layer Profiling}
%\title{Inference-Time Rebalance via Layered Modality Equilibrium and Propagation for Multimodal Emotion Reasoning}
%\title{Layer Roles Matter: Layer-Aware Modality Rebalance in Multimodal Emotion Reasoning}
%\title{Layer Roles Matter: Layer-Aware Modality Equilibrium and Propagation in Multimodal Emotion Reasoning}


\author{Pengyu Huang}
\affiliation{%
  \institution{University of Science and Technology of China}
  \city{Hefei}
  \country{China}
}
\email{otishuang@163.com}

\author{Peipei Song}
\authornote{Corresponding authors.}
\affiliation{%
  \institution{University of Science and Technology of China}
  \city{Hefei}
  \country{China}
}
\email{beta.songpp@gmail.com}

\author{Beier Zhu}
\authornotemark[1]
\affiliation{%
  \institution{University of Science and Technology of China}
  \city{Hefei}
  \country{China}
}
\email{beier.zhu@ustc.edu.cn}

\author{Zhiyuan Han}
\affiliation{%
  \institution{University of Science and Technology of China}
  \city{Hefei}
  \country{China}
}
\email{aaronhan@mail.ustc.edu.cn}

\author{Jinpeng Hu}
\affiliation{%
  \institution{Hefei University of Technology}
  \city{Hefei}
  \country{China}
}
\email{135858hjp@gmail.com}

\author{Xun Yang}
\authornotemark[1]
\affiliation{%
  \institution{University of Science and Technology of China}
  \city{Hefei}
  \country{China}
}
\email{xyang21@ustc.edu.cn}

\renewcommand{\shortauthors}{Huang et al.}

\begin{abstract}
Multimodal Emotion Reasoning (MER) requires perceiving and integrating emotion cues from multiple modalities. Recent advances increasingly leverage Multimodal Large Language Models (MLLMs) to enhance reasoning capacity. However, due to the inherent heterogeneity in conveying emotions across modalities, these models often suffer from modality imbalance and confront flawed emotion understanding. In this paper, we conduct an in-depth analysis on MER models and uncover two key empirical insights. (\textit{i}) In MER, modality imbalance manifests two distinct failure modes---unimodal over-reliance and insufficient cross-modal utilization, which call for different intervention strategies. (\textit{ii}) MLLM layers exhibit implicit functional hierarchies in multimodal emotion processing, playing separable roles in stabilizing unimodal representations and integrating cross-modal information.
Motivated by these observations, we propose a \texttt{L}ayer-\texttt{A}ware \texttt{M}odality \texttt{R}ebalancing (\texttt{LAMR}) framework by explicitly aligning intervention strategies with layer roles.
Specifically, we first introduce two quantized metrics for adaptive layer role profiling. In each layer, Modality Entropy (ME) measures the dispersion of modality contributions, and Cross-modal Intensity (CMI) captures the strength of cross-modal interactions. Then, we respectively enhance modality equilibrium in stabilization layers and cross-modal propagation in integration layers.
Extensive experiments on two comprehensive MER benchmarks demonstrate that \texttt{LAMR} effectively mitigates modality imbalance and improves emotion reasoning across diverse MLLM backbones, without additional training. Project url: \url{https://pengyuhuang1.github.io/LAMR/}
\end{abstract}



\begin{teaserfigure}
    \centering
    \includegraphics[width=0.96\textwidth]{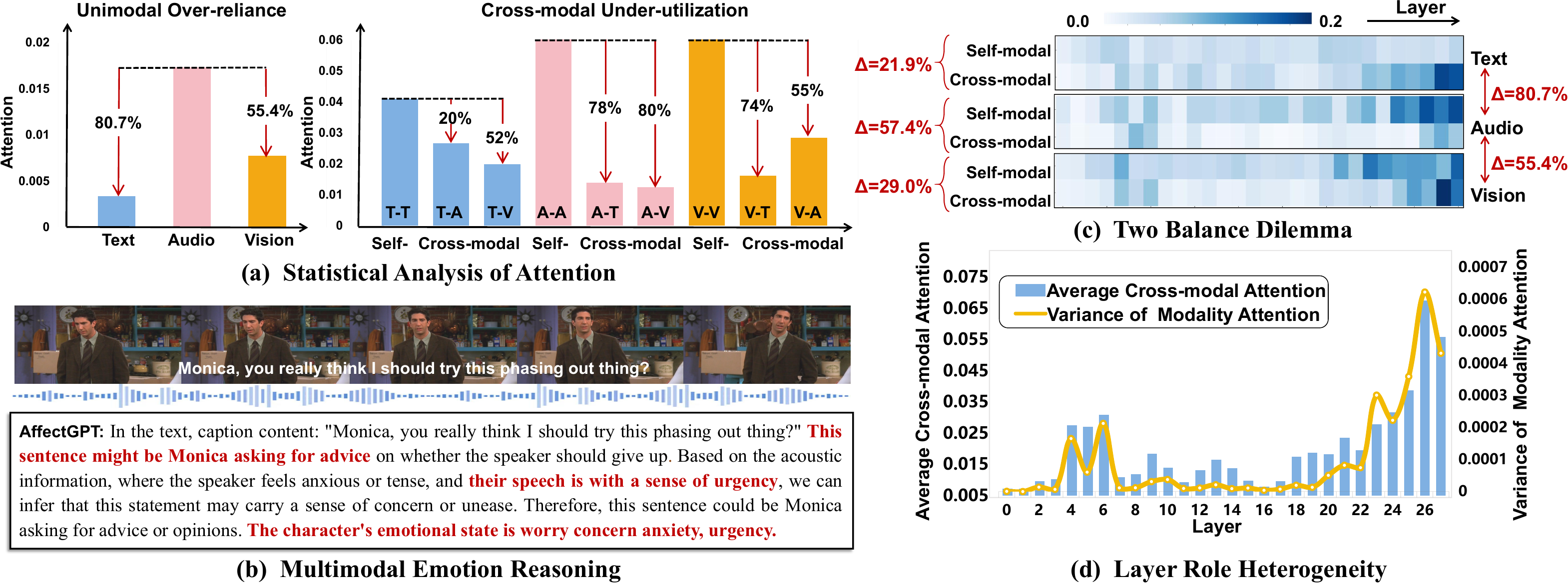}
    \caption{Diagnostic analysis of attention imbalance and layer-wise heterogeneity in MLLM-based emotion reasoning. \textbf{(a)} Attention statistics of AffectGPT. 
    % (see Appendix~\ref{appendix:panda&emo} for PandaGPT and EmotionLLaMA).
    \textbf{(b)} An example of erroneous emotion reasoning given by AffectGPT. \textbf{(c)} Layer-wise self-modal and cross-modal attention heatmaps. \textbf{(d)} Layer-wise evolution of modality attention gaps and cross-modal attention.
}
    \Description{Four panels diagnose modality imbalance in AffectGPT: modality-level attention statistics, an erroneous multimodal emotion reasoning example, layer-wise self- and cross-modal attention heatmaps, and layer-wise trends in cross-modal attention and modality imbalance.}
    \label{fig:introduction}
\end{teaserfigure}

\maketitle

\section{Introduction}
\label{sec:intro}
%任务重要性 + MLLMs 的机会
Multimodal Emotion Reasoning (MER)~\cite{cheng2024emotion-llama,lian2023emer,huang2025emotion-qwen,Bridging_Subjectivity,han2026omniperception} is a challenging task that requires models not only to extract emotion-related cues from individual modalities, but also to perform deep reasoning over multimodal interactions to capture complex emotion expressions and contexts. With the emergence of Multimodal Large Language Models (MLLMs) \cite{su2023pandagpt}, MER has shifted toward a unified generative reasoning paradigm. 
%the paradigm for solving MER has undergone substantial changes. 
Recent approaches, such as Emotion-LLaMA \cite{cheng2024emotion-llama}, AffectGPT \cite{Lian2025AffectGPT}, and Emotion-Qwen \cite{huang2025emotion-qwen}, leverage MLLMs to jointly process and integrate multimodal data, achieving notable improvements in emotion recognition and reasoning.

%MER中的模态不平衡问题
In MER, different modalities are often complementary, providing additional cues for semantic and emotional disambiguation. However, due to inherent and complex differences in how modalities capture and express information, as well as the varying informational richness of each modality, the MER models inevitably face modality balance dilemma~\cite{mullick2025text,zhang2025moda}. This implies two ensuing issues, as highlighted in the statistical analysis of AffectGPT (Figure~\ref{fig:introduction} (a)): \textbf{unimodal over-reliance}, where models tend to favor a dominant modality; and \textbf{cross-modal under-utilization}, where self-modal and cross-modal interactions are severely imbalanced.
These imbalance issues ultimately result in flawed emotion understanding. As shown in Figure~\ref{fig:introduction} (b), audio dominance induces incorrect recognition of \textit{urgency}, while text-only reasoning causes context confusion (\textit{e.g., Monica asking for advice}).

%解决模态不平衡的现有方法，瓶颈：训练成本，只关注一个问题，层次
Recently, some efforts have attempted to address the modality balance dilemma in multimodal emotion understanding. EMOE~\cite{fang2025emoe} proposes dynamic modality weighting to prevent over-reliance on dominant modalities during training. MoSEAR~\cite{han2025benchmarking} employs modality-specific experts with a regularized gating mechanism in fine-tuning heads. MODA~\cite{zhang2025moda} further tackles attention deficit disorder through modular duplex attention. %decomposes attention into self-modal and cross-modal parts. 
However, the additional fine-tuning costs in these methods limit their scalability and practicality in applications. Training-free methods such as visual augmentation~\cite{fangcatch} and contrastive decoding~\cite {jung2025avcd} have been explored to mitigate modality bias. %to address the modality imbalance using approaches . %Training-free methods have attracted increasing interest. SEPM~\cite{fangcatch} enhances emotion perception by directing the modal attention to emotion-relevant regions. AVCD~\cite{jung2025avcd} applies attentive masking to encourage more balanced multimodal reasoning.
Although promising, existing methods still exhibit crucial limitations. 
(1) they primarily focus on alleviating unimodal dominance, while overlooking the distinct challenge of the imbalance between self-modal and cross-modal interactions;
(2) they typically adopt layer-agnostic strategies, which is suboptimal given the functional heterogeneity of multimodal processing across layers~\cite{jiang2025devils,zhang2025cross,guo2025lisa}. 

To address the above limitations, we delve into how MLLM-based MER models process multimodal tokens and how this affects the generation of modality imbalance. 
%Our analysis leads to two key observations. 
As shown in Figure~\ref{fig:introduction} (c), the self- and cross-modal attention heatmaps across modalities reveal two coexisting failure modes of modality imbalance. Figure~\ref{fig:introduction} (d) further shows a clear layer-wise dependency, such as layers with small modality gaps consistently exhibit weak cross-modal interactions.
These observations underscore the need for rebalance strategies that explicitly account for the two-sided attention dilemma and layer-wise functional roles.

In this paper, we propose \texttt{L}ayer-\texttt{A}ware \texttt{M}odality \texttt{R}ebalancing (\texttt{LAMR}), which specifies intervention strategies aligned with layer roles. \textbf{First}, we introduce an adaptive layer role profiling strategy that characterizes the evolution of multimodal attention across layers. Instead of a heuristic solution, we propose two metrics of Modality Entropy (ME) and Cross-Modal Intensity (CMI), which respectively quantify the balance of modality participation and the strength of cross-modal interactions at each layer. These metrics are combined to identify a functional bifurcation across layers and partition the MLLM into stabilization and integration layers, where the former primarily stabilizes modality-specific representations, and the latter facilitates cross-modal semantic integration.
\textbf{Second}, we propose layer-wise attention re-balancing tailored to the identified layer roles. In stabilization layers, we enforce \textit{modality equilibrium} to suppress abnormal unimodal dominance and stabilize attention flow. In integration layers, we control \textit{emotion propagation} via a cross-modal affinity graph, facilitating effective diffusion of semantically relevant information across modalities while preserving the original representational structure. 
\texttt{LAMR} is fully plug-and-play and can be seamlessly integrated into existing MLLM-based MER models, scaling the emotion intelligence during the inference stage. 
Our contributions are summarized as follows:
% \vspace{-2mm}
\begin{itemize}
\setlength{\itemsep}{0pt}
\setlength{\parsep}{0pt}
\setlength{\parskip}{0pt}
    \item We identify a two-sided attention dilemma in MLLM-based emotion reasoning---unimodal over-reliance and cross-modal under-utilization---and reveal its intrinsic connection to layer-wise functional heterogeneity.
    \item We propose \texttt{LAMR} for MER, a novel inference-time framework that automatically profiles layer roles via two quantitative metrics and performs role-conditioned attention rebalancing. 
    \item Extensive experiments on different MER benchmarks and MLLM backbones demonstrate that \texttt{LAMR} effectively mitigates modality imbalance and improves emotion recognition and reasoning without any retraining. 
\end{itemize}

%%%%%%%%%%%%%%%%%%%%%%%%%%%%%%%%%%%%%%%%%%%%%%%%%%%%%%%%%%%%%%
\section{Related Work}
\label{sec:related}
\subsection{Multimodal Emotion Reasoning}
{Multimodal emotion reasoning} requires joint reasoning over visual, linguistic, and acoustic cues. The recent emergence of MLLMs has advanced MER from early single-label emotion recognition~\cite{yang2024emollmmultimodalemotionalunderstanding,zhang2026benchmarkingdynamic,huBeyondEmotion,SongEmotional_Video} to open-vocabulary and explainability-oriented frameworks~\cite{lian2023emer,lian2025ovmer}. %However, general MLLMs still struggle to perform reliable emotion reasoning without task-specific adaptation \cite{su2023pandagpt}[??]. To this end, 
Building on generalist MLLMs~\cite{su2023pandagpt}, a line of emotion-specialized models has emerged through post-training adaptation, including Emotion-LLaMA~\cite{cheng2024emotion-llama}, R1-Omni~\cite{r1omni}, AffectGPT~\cite{Lian2025AffectGPT}, and Emotion-Qwen~\cite{huang2025emotion-qwen}. Beyond affective computing, modal joint learning has also been explored in unified autonomous driving systems such as MOJITO~\cite{cheng2026mojitomodaljointlearning}.
% Emotion-LLaMA projects visual, audio, and textual features into a unified LLM space via emotion-specific encoders, followed by instruction tuning to activate emotional reasoning capabilities.R1-Omni enhances emotional intelligence through reinforcement learning with verified feedback (RLVR), while AffectGPT leverages large-scale emotional datasets and a pre-fusion projector to strengthen multimodal emotion integration.Emotion-Qwen further explores emotion-aware multimodal alignment within the Qwen family of MLLMs.
In parallel, numerous efforts have been made to establish standardized benchmarks for evaluating multimodal emotional intelligence. Recent benchmarks such as EmoBench~\cite{sabour-etal-2024-emobench}, MER-UniBench~\cite{Lian2025AffectGPT}, and MME-Emotion~\cite{zhang2025mme-emotion} provide holistic and rigorous analytical evaluation protocols and reveal the great potential in multimodal emotion reasoning.

\begin{figure*}[t]
    \centering
    \includegraphics[width=1\textwidth]{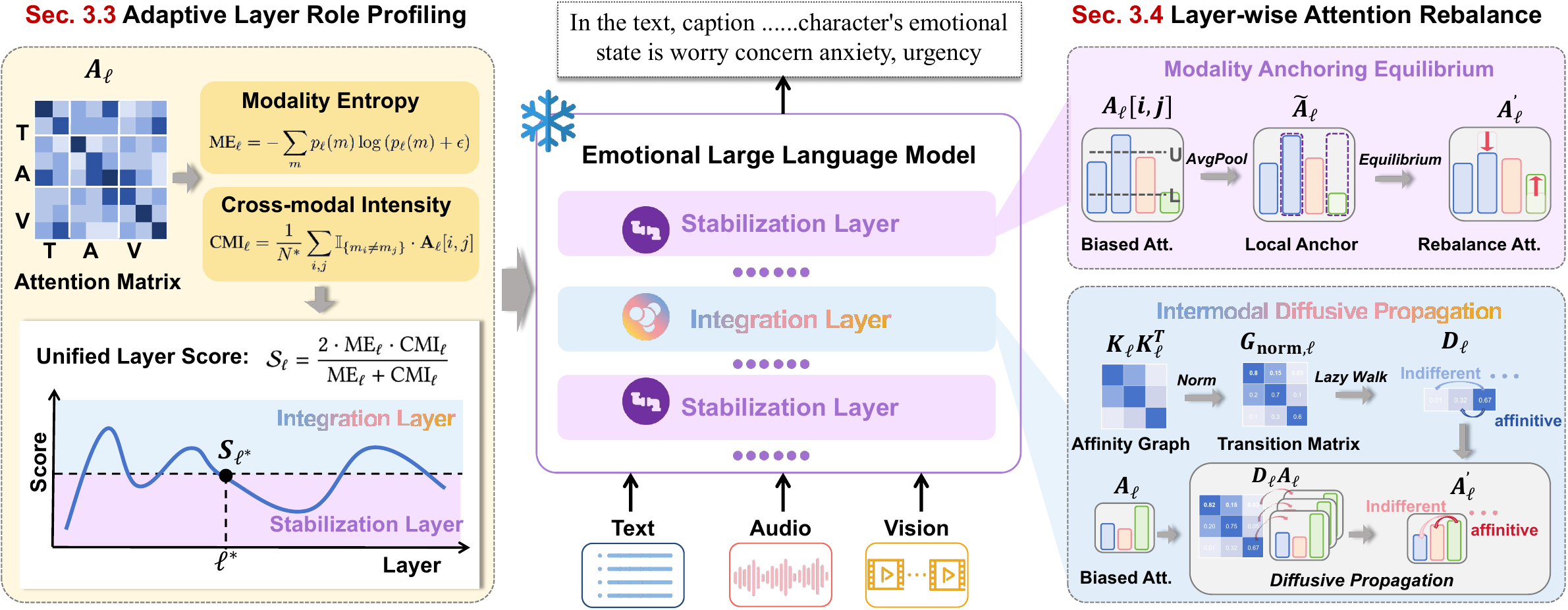}
    \caption{Overview of the proposed \texttt{LAMR} framework. \textit{Adaptive Layer Role Profiling} automatically partitions Transformer layers into stabilization layers and integration layers. \textit{Layer-wise Attention Rebalance} performs layer-specific attention interventions during inference, mitigating unimodal over-reliance in stabilization layers and enhancing cross-modal propagation in integration layers.
% \textbf{(a)} \textit{Adaptive Layer Role Profiling} identifies stabilization layers and integration layers in an unsupervised manner. 
% \textbf{(b)} \textit{Layer-wise Attention Rebalance} implements layer-specific intervention strategies during inference. By decoupling intra-modal semantic preservation from inter-modal interaction, our approach mitigates the aforementioned issues of Uni-modal Over-reliance and Cross-modal Under-utilization.
}
    \Description{Pipeline of LAMR. Layer-wise modality entropy and cross-modal intensity are combined to profile stabilization and integration layers, after which modality equilibrium and cross-modal propagation are applied according to the identified layer roles.}
    \label{fig:methohd}
\end{figure*}

\subsection{Modality Imbalance}
{Modality imbalance} has emerged as a critical challenge in multimodal learning. In the context of general MLLMs, several studies show that imbalanced modality utilization often induces hallucinations and unreliable reasoning~\cite{zhang2025robust,zhou2024mitigating,jung2025avcd}. 
% MMMC~\cite{zhang2025robust} addresses robustness against modality conflict by encouraging consistency across modalities, while CausalMM~\cite{zhou2024mitigating} mitigates modality-prior–induced hallucinations by analyzing attention causality and disentangling spurious modality dominance.  to suppress hallucinations by selectively perturbing over-confident modalities at inference time.
For instance, AVCD~\cite{jung2025avcd} suppresses modality-induced hallucinations in audio-visual LLMs by contrastive decoding.
In MER, this issue becomes more pronounced due to the subjective and context-dependent nature of emotional expressions~\cite{mullick2025text,fang2025emoe,fangcatch}. Recent efforts such as MoSEAR~\cite{han2025benchmarking} explicitly benchmark emotion conflicts across modalities and introduce expert-based mechanisms to alleviate modality bias in multimodal emotion reasoning. MODA~\cite{zhang2025moda} further proposes modular duplex attention that disentangles self-modal and cross-modal attention to improve multimodal perception and emotion understanding. Despite their effectiveness, existing approaches typically rely on additional fine-tuning and overlook layer-wise functional heterogeneity in MLLMs. In contrast, we aim to improve the emotional reasoning ability of MLLMs solely during inference by performing layer-wise rebalancing. 

\subsection{Layer-wise Interpretability of MLLMs}
{Layer-wise interpretability of MLLMs} has attracted a great deal of attention in the community. DoLa~\cite{chuang2024dola} reveals that factual knowledge in LLMs is generally localized to particular transformer layers.
% reveals that factual knowledge is hierarchically encoded across transformer layers and exploits layer contrasting to improve decoding factuality. 
Neo et al.~\cite{neo2025towards} study the processing of visual tokens in the vision-language model and observe their increasing interpretability in the vocabulary space across layers. 
%find that visual token representations become progressively interpretable across layers in VLMs, suggesting a gradual alignment between visual and textual modalities.} 
CMIF~\cite{zhang2025cross} further elucidates the cross-layer information flow between different modalities. Building on these analytical insights, LISA~\cite{guo2025lisa} leverages the functional roles of transformer layers to mitigate object hallucinations. Devils~\cite{jiang2025devils} identifies that middle layers disproportionately contribute to hallucination in Large Vision-Language Models (LVLMs) and adjusts visual attention accordingly. Yue et al.~\cite{yang2025understanding} argue that hallucination attention heads are concentrated in the middle and deeper layers, displaying a strong attention bias toward text tokens. They propose two methods that are targeted for hallucination heads to reduce their reliance on text tokens. 
%discover that hallucination-prone attention heads concentrate in middle-to-deep layers and propose targeted head-level interventions.} 
Inspired by these findings, we focus on developing a layer-wise intervention strategy tailored to multimodal emotion reasoning.

%%%%%%%%%%%%%%%%%%%%%%%%%%%%%%%%%%%%%%%%%%%%%%%%%%%%%%%%%%%%%%
\section{Methodology}

\label{sec:method}
% {\color{blue}Overview of the method, cite Figure 2,... or delete}
% {\color{blue}
% In this section, we first outline the preliminary formulation and analyze empirical interaction patterns to highlight the modality imbalance. Next, we introduce Adaptive Layer Role Profiling metrics to identify critical integration layers. Finally, we detail Layer-wise Attention Rebalance, employing modality anchoring and intermodal diffusion mechanisms to optimize multimodal fusion.
% }

\subsection{Preliminary}
\label{subsec:preliminary}
We formulate MER within the framework of MLLMs \cite{cheng2024emotion-llama}, where visual, acoustic, and textual inputs are jointly represented. Each modality is first encoded into the MLLM embedding space to form modality-specific token sequences $\mathbf{T}_m = \mathrm{Enc}_m(\mathbf{x}_m) \in \mathbb{R}^{N_m \times d}$, where $m \in \{{v}, {a}, {t}\}$. The initial input is constructed by concatenation: $\mathbf{X}_0 = [\,\mathbf{T}_{t};\ \mathbf{T}_{v};\ \mathbf{T}_{a}\,] \in \mathbb{R}^{N \times d}$,
where $N = N_t + N_v + N_a$ denotes the total sequence length.
The MLLM then processes $\mathbf{X}_0$ through a stack of Transformer layers. At the $\ell$-th layer, the input $\mathbf{X}_{\ell-1}$ is projected into queries, keys, and values: $\{\mathbf{Q}_\ell, \mathbf{K}_\ell, \mathbf{V}_\ell\} = \mathbf{X}_{\ell-1} \mathbf{W}^{\{Q,K,V\}}_\ell$. The attention matrix and layer output are defined as:
\begin{equation}
    \mathbf{A}_\ell = \mathrm{softmax}( \frac{\mathbf{Q}_\ell \mathbf{K}_\ell^\top}{\sqrt{d_h}}), \quad \mathbf{X}_\ell = \mathbf{A}_\ell \mathbf{V}_\ell,
\end{equation}
where $\mathbf{A}_\ell \in \mathbb{R}^{N \times N}$ encodes both \emph{self-modal} and \emph{cross-modal} interactions, which play different roles in emotion reasoning. For elements $i,j$ with modality labels $m_i,m_j$, entries with $m_i = m_j$ correspond to {self-modal} attention, which primarily consolidates intra-modal emotional cues (\eg, prosodic patterns in audio or facial dynamics in vision), contributing to modality-specific representation stabilization. In contrast, entries with $m_i \neq m_j$ form {cross-modal} attention, which enables the exchange and alignment of complementary evidence across modalities, facilitating multimodal integration. 
In this work, we focus on analyzing and refining the attention matrix $\mathbf{A}_\ell$, as it explicitly governs how multimodal evidence is weighted and propagated across tokens.

% \noindent \textbf{Self-modal and Cross-modal Attention.}
% Based on the attention computation defined above, the mean attention score from source modality $m$ to target modality $n$ (where $m, n \in \{v, a, t\}$) is calculated as:
% We quantify the mean attention from source modality $m$ to target modality $n$ at layer $\ell$ as
% \begin{equation}
%     \mathcal{A}^{m \to n}_{\ell} = \frac{1}{N_m N_n} \sum_{i \in \mathcal{I}^m} \sum_{j \in \mathcal{I}^n} \mathbf{A}_{\ell}[i, j],
% \end{equation}
% where $\mathcal{I}^m$ denotes the index set of tokens of modality $m$. When $m = n$, $\mathcal{A}^{m \rightarrow n}_{\ell}$ corresponds to self-modal attention, and when $m \neq n$, it describes cross-modal attention. %When $m = n$, $\mathcal{A}^{m \rightarrow n}_{\ell}$ corresponds to self-modal attention, and when $m \neq n$, it characterizes cross-modal attention. and $\mathbf{A}_{\ell}[i, j]$ refers to the attention score from token $i$ to token $j$. Formally, $\mathcal{A}^{m \to n}_{\ell}$ denotes the self-modal attention when $m=n$, and the corresponding cross-modal attention when $m \neq n$.

%问题提出
\subsection{Multimodal Attention Diagnostics}
\label{subsec: observation}
% \noindent \textbf{Experimental setup.}
% To understand how multimodal emotional evidence is processed during inference, we investigate the layer-wise attention patterns in three representative MER models, including AffectGPT~\cite{Lian2025AffectGPT}, PandaGPT~\cite{su2023pandagpt}, and EmotionLLaMA~\cite{cheng2024emotion-llama}, using the CMU-MOSEI and CMU-MOSI datasets ~\cite{bagher-zadeh-etal-2018-multimodal}. We observe consistent trends across models; for clarity, the analysis on AffectGPT is presented in the main text, and others are included in Appendix~\ref{appendix:panda&emo}.
% Unlike general multimodal tasks where a dominant modality often suffices, 
Compared with general multimodal tasks, emotion reasoning fundamentally relies on the subtle interplay among modalities. For example, a speaker's words may convey politeness, while their tone reveals frustration. This characteristic makes MER particularly susceptible to modality imbalance, yet it remains unclear how such imbalance emerges within MLLM-based MER models. 
To shed light on this issue, we investigate the layer-wise attention patterns in three representative MER models, including AffectGPT~\cite{Lian2025AffectGPT}, PandaGPT~\cite{su2023pandagpt}, and EmotionLLaMA~\cite{cheng2024emotion-llama}, using the CMU-MOSEI and CMU-MOSI datasets ~\cite{bagher-zadeh-etal-2018-multimodal}. The attention matrix $\mathbf{A}_\ell$ provides a natural diagnostic lens to examine how multimodal emotional evidence is processed during inference. 
% We observe consistent trends across models; for clarity, the analysis on AffectGPT is presented in the main text, and others are included in Appendix~\ref{appendix:panda&emo}.

\noindent \textit{\textbf{Observation 1: Two kinds of balance dilemma.}}
As shown in Figure~\ref{fig:introduction} (c), our analysis reveals two co-existing but conceptually distinct imbalance phenomena. First, attention primarily concentrates on a dominant modality (such as audio), marginalizing complementary cues from other modalities. Second, even when multiple modalities are attended to, the majority of attention remains confined within self-modal interactions, while cross-modal attention remains sporadic. %Importantly, these two issues do not collapse into a single form of imbalance: suppressing unimodal dominance alone does not necessarily induce effective cross-modal interaction, nor does amplifying cross-modal attention guarantee balanced modality participation. 
This observation suggests that {unimodal over-reliance} and {cross-modal under-utilization} are two orthogonal bottlenecks in multimodal emotion reasoning, each imposing distinct constraints on attention regulation.

\noindent \textit{\textbf{Observation 2: Layered functional heterogeneity.}}
Beyond global imbalance, we observe that these attention behaviors are not uniformly distributed across Transformer layers. As illustrated in Figure~\ref{fig:introduction} (d), layers with small modality-wise attention disparities tend to exhibit weak cross-modal interactions, whereas layers with pronounced unimodal dominance often coincide with stronger cross-modal attention. 
This layer-wise divergence indicates that different Transformer layers are implicitly specialized for distinct functional roles during emotion inference. Specifically, layers with balanced but weak interactions primarily serve to {stabilize} modality-specific representations, whereas layers with pronounced cross-modal attention focus on {integrating} complementary emotional cues across modalities. 
This functional bifurcation suggests that Transformer layers can be naturally partitioned into two categories, each calling for a distinct intervention strategy. We will formalize this principle in the following sections.

% intra-modal attention consistently dominates in early layers, while cross-modal interactions remain sparse and fragmented. Notably, salient cross-modal attention only emerges in a subset of deeper layers, indicating that semantic integration is not a pervasive process but a selective one confined to specific stages of inference. Consequently, treating all layers identically when regulating attention—either by enforcing uniform modality balance or amplifying cross-modal interaction—can be suboptimal or even detrimental. These findings reveal a distinct layered heterogeneity, motivating the need for quantitative metrics to characterize these layer-specific roles, which we introduce in the following section.

%\textbf{Adaptive Layer Role Profiling Metrics.}  \textbf{Left:} Layer-wise Cross-Modal Intensity (CMI) and Modality Entropy (ME) on AffectGPT. 

\subsection{Adaptive Layer Role Profiling}
\label{subsec:adaptive_layer}
Motivated by the aforementioned findings, we propose an \textit{adaptive layer role profiling} strategy to characterize the functional roles of Transformer layers in multimodal emotion reasoning. Instead of predefining layer roles heuristically, our approach includes two intrinsic metrics derived from attention patterns---Modality Entropy (ME) and Cross-Modal Intensity (CMI)---to identify whether a layer primarily serves to stabilize unimodal evidence or actively integrate cross-modal information. %ME reflects the breadth of modality participation within a layer, while CMI captures the strength of cross-modal attention.  
% : 

\subsubsection{\textit{\textbf{Modality Entropy}}} 
ME is defined based on Shannon entropy~\cite{Shannon} to quantify the balance of attention allocation across modalities. At the $\ell$-th layer, we obtain:
\begin{equation}
    \label{eq:me}
\begin{cases}\text{ME}_{\ell} = - \sum_{m} p_{\ell}(m) \log \left( p_{\ell}(m) + \epsilon \right),\\
p_{\ell}(m) = \frac{1}{N^*}  \sum_{i,j}\mathbb{I}_{\{ m_j = m\}} \cdot\mathbf{A}_{\ell}[i, j], \end{cases}
\end{equation}
where $p_{\ell}(m)$ represents the probability mass of attention allocated to modality $m$ at layer $\ell$. $\mathbb{I}_{\{\cdot\}}$ is the indicator function, $m \in \{{v}, {a}, {t}\}$ denotes the modality of token, and $N^*$ is the normalization constant.
%$\epsilon$ is a..., , and . Higher $\text{ME}_{\ell}$ indicates that multiple modalities are comparably engaged. 
Given that entropy is maximized when probability mass is evenly distributed and minimized when concentrated on a single component, ME provides a principled measure of unimodal dominance. A higher $\text{ME}_{\ell}$ indicates that attention is more evenly distributed across modalities, whereas a lower value reflects strong reliance on a dominant modality.
%两个值都高，算交互层，仅ME高或仅CMI高，算支撑层

\subsubsection{\textit{\textbf{Cross-modal Intensity}}}
In $\mathbf{A}_{\ell}$, cross-modal attention enables the exchange and integration of complementary emotional cues between modalities, whereas self-modal attention refines relationships within each modality. Focusing on cross-modal flow, CMI quantifies cross-modal attention intensity at the $\ell$-th layer:
\begin{equation}
\label{eq:CMI}
    \text{CMI}_{\ell} = \frac{1}{N^*} \sum_{i,j}\mathbb{I}_{\{m_i \neq m_j\}} \cdot \mathbf{A}_{\ell}[i, j].
    % \text{CMI}_{\ell} = \frac{1}{N} \sum_{i,j} \mathbb{1}_{\{m_i \neq m_j\}} \cdot \mathbf{A}_{\ell}[i, j],
\end{equation}
Since $\mathbf{A}_{\ell}[i, j]$ is normalized by applying softmax given both cross-modal and self-modal pairs, %$\text{CMI}_{\ell}$ represents the probability that the layer will not contain any self-modal.
$\text{CMI}_{\ell}$ increases with an increase in cross-modal scores relative to the self-modal scores, suggesting that attention is actively propagated across modalities.

ME and CMI capture two complementary aspects of multimodal attention. However, there is a pitfall: a layer may exhibit high CMI by repeatedly propagating attention from a single modality to others; conversely, a layer with high ME may distribute attention evenly across modalities but lack meaningful cross-modal interactions. Effective multimodal reasoning requires both balanced modality engagement and active cross-modal integration. To this end, we compute the unified layer score $\mathcal{S}_{\ell}$ as the harmonic mean~\cite{harmonic} of ME and CMI. This formulation requires both metrics to be elevated for a high score: 
\begin{equation}
\label{eq:sl}
    \mathcal{S}_{\ell} = \frac{2 \cdot \text{ME}_{\ell} \cdot \text{CMI}_{\ell}}{\text{ME}_{\ell} + \text{CMI}_{\ell}}.
\end{equation}
To adaptively identify the transition boundary $\ell^*$, we apply the Kneedle algorithm~\cite{satopaa2011knee} to the set of scores $\mathcal{S}_{\ell}$ sorted in descending order. This algorithm mathematically locates the point of maximum curvature, allowing us to pinpoint the precise layer where functional roles diverge without relying on rigid manual heuristics.

Intuitively, a higher $\mathcal{S}_{\ell}$ indicates that the layer concurrently supports active cross-modal interaction and balanced modality participation, making it suitable for semantic integration, whereas other layers primarily function to stabilize modality-specific representations.
Therefore, we designate layers with $\mathcal{S}_{\ell} \geq \mathcal{S}_{\ell^*}$ as \textbf{integration layers}, and layers with lower $\mathcal{S}_{\ell}$ as \textbf{stabilization layers}.

% \begin{figure}[t] % h=here, t=top, b=bottom, p=page
%     \centering
%     \includegraphics[width=\columnwidth]{Figs/figure5.png}
%     \caption{Layers are ranked by $\mathcal{S}_{\ell}$ to identify the transition between stabilization and integration layers. %\color{blue}Layers ranked by the unified layer score ($\mathcal{S}_{\ell}$)
%     }
%     \label{fig:layerperception1}
% \end{figure}

%理论加强
\subsection{Layer-wise Attention Rebalance}
\label{subsec: overall method}
In accordance with the distinct functional roles of stabilization and integration layers, we design a layer-specific rebalancing strategy that addresses unimodal over-utilization in stabilization layers and cross-modal under-utilization in integration layers, respectively.

\subsubsection{\textit{\textbf{Stabilization Layer: Modality Anchoring Equilibrium}}}
\label{subsubsec:modality_anchoring}
To establish a reliable multimodal evidence basis, % before higher-level semantic integration, 
we aim to anchor attention distributions across modalities into a balanced regime.
Rather than relying on absolute attention magnitudes, we construct a layer-wise reference to characterize typical attention intensity. Given $\mathbf{A}_\ell$, where each row represents a token-wise attention distribution, we compute a local anchor by average pooling along the row dimension $\tilde{\mathbf{A}}_\ell = \text{AvgPool}(\mathbf{A}_\ell)$. %{\color{blue} Formally, $\text{AvgPool}$ marginalizes over token-wise distributions to smooth local fluctuations, yielding a robust statistical baseline for layer-wise intensity calibration.}
This anchor $\tilde{\mathbf{A}}_\ell$ captures the expected attention scale of layer $\ell$ and provides a modality-agnostic anchor against token-level fluctuations.
% Here, each row of the attention map represents the independent attention distribution of a specific token over the global context. The $\text{AvgPool}$ operation smooths high-frequency local fluctuations and extracts a statistical benchmark reflecting the general intensity of the current attention map, thereby avoiding misjudgments caused by relying solely on absolute numerical values.

Importantly, unimodal over-reliance manifests not as uniformly elevated attention, but as extreme deviations concentrated on a subset of tokens or modalities. To distinguish pathological dominance from legitimate salient cues, we leverage the Interquartile Range (IQR) method~\cite{Shang_2025_ICCV, IQR}, a statistical technique known for its robustness in outlier detection. Specifically, we define an equilibrium interval $[L, U]$ that delineates the normal range of attention values here:
\begin{equation}
\label{eq:iqr}
    L = Q_1 - 1.5\times{IQR}, \quad U = Q_3 + 1.5\times{IQR},
\end{equation}
where $Q_1$ and $Q_3$ denote the first and third quartiles of $\mathbf{A}_\ell$, and $IQR=Q_3-Q_1$. Attention values outside $[L, U]$ are considered indicative of abnormal confidence concentration.

Finally, outlier attentions are shrunk toward the local anchor $\tilde{\mathbf{A}}_\ell$:
\begin{equation}
\label{eq:mae}
\mathbf{A}'_\ell = \begin{cases} (1 - \alpha_s) \mathbf{A}_\ell \pm \alpha_s \tilde{\mathbf{A}}_\ell, & \mathbf{A}_\ell \notin [L, U], \\ \mathbf{A}_\ell, & \text{otherwise}. \end{cases}
\end{equation}
where $\alpha_s$ controls the anchoring strength. %Instead of hard clipping, which would discard information and potentially disrupt inference dynamics, this soft regularization contracts the variance of attention distributions while preserving their relative structure, thereby mitigating unimodal dominance without enforcing artificial uniformity.
Instead of hard clipping, this soft regularization contracts attention variance while preserving relative structure, thereby mitigating unimodal dominance without enforcing uniformity.

\subsubsection{\textit{\textbf{Integration Layer: Intermodal Diffusive Propagation}}}
\label{subsubsec:intermodal_diffusion}
Emotion reasoning requires that semantically aligned cues from different modalities reinforce each other. This motivates a content-driven substrate to model latent semantic correlations for information diffusion.
%{\color{blue}Such reinforcement should be guided by intrinsic semantic correlations, instead of query-specific attention decisions that may already be biased. }
As shown in Figure \ref{fig:methohd}, we first construct a semantic affinity graph based on the key representations:
\begin{equation}
\label{eq:gram}
    \mathbf{G}_\ell = \mathbf{K}_\ell \mathbf{K}_\ell^\top \in \mathbb{R}^{N \times N}.
\end{equation}
% where each entry $\mathbf{G}_\ell(i,j)$ measures the semantic similarity between tokens $i$ and $j$.
Unlike raw attention scores, the key-based affinity captures symmetric and current query-independent associations. However, $\mathbf{G}_\ell$ typically reflects strong self-token similarity. To enable cross-modal information transfer, we perform row-wise normalization on $\mathbf{G}_\ell$, which redistributes the self-affinity toward semantically related cross-modal tokens and yields a transition matrix $\mathbf{G}_{\mathrm{norm},\ell}$.
% {\color{blue}
% \begin{equation}
% \mathbf{G}_{\mathrm{norm},\ell}(i,j) = \frac{\mathbf{G}_\ell(i,j)}{\sum_{j \in \mathcal{N}}\mathbf{G}_\ell(i,j) + \xi},
% \end{equation}
% where $\xi = 10^{-6}$ ensures numerical stability.
% }

% To enable information propagation over this semantic graph, 
% \begin{equation}
%     \mathbf{G}_{\mathrm{norm},\ell}(i,j)
%     = \frac{\mathbf{G}_\ell(i,j)}{\sum_{j \in \mathcal{N}_{\text{cross}}}\mathbf{G}_\ell(i,j) + \xi},
% \end{equation}
% where $\xi = 10^{-6}$ ensures numerical stability. 

Then, we define a diffusive propagation operator that corresponds to a lazy random walk on $\mathbf{G}_{\mathrm{norm},\ell}$:
\begin{equation}
\label{eq:diffusive operater}
    \mathbf{D}_\ell = (1 - \alpha_d)\mathbf{I} + \alpha_d\,\mathbf{G}_{\mathrm{norm},\ell},
\end{equation}
where $\alpha_d$ controls the diffusion strength, and $\mathbf{I}$ is the identity matrix. %to preserve the original attention structure. %and the diffusion term enables controlled propagation of attention mass across semantically aligned tokens. 
% From this perspective, attention refinement is recast as a diffusion process over a semantic graph defined in the key space.
Since $\mathbf{D}_\ell$ is row-stochastic, it preserves the row-wise normalization of attention.
The biased attention $\mathbf{A}_\ell$ is thus refined by applying the diffusion operator
% \begin{equation}
    $\mathbf{A}'_\ell = \mathbf{D}_\ell \mathbf{A}_\ell$.
% \end{equation}
For each query token $i$, it yields:
\begin{equation}
\begin{aligned}
\label{eq:idp}
\mathbf{A}'_\ell(i,:)
= &(1 - \alpha_d)\mathbf{A}_\ell(i,:)\\
&+ \alpha_d \sum_{j}\mathbf{G}_{\mathrm{norm},\ell}(i,j)\,\mathbf{A}_\ell(j,:).
\end{aligned}
\end{equation}
That is, each query keeps a fraction $(1-\alpha_d)$ of its original attention while absorbing an $\alpha_d$-weighted mixture of attention patterns from semantically related cross-modal neighbors, thereby alleviating cross-modal under-utilization in emotion reasoning.

\begin{table*}[t]
\centering
\scriptsize
\caption{Comparison on MER-UniBench \cite{Lian2025AffectGPT} for multimodal emotion recognition. This benchmark comprises nine datasets covering three tasks: basic emotion recognition, sentiment analysis, and fine-grained emotion recognition. %We apply our proposed method to three emotion-aware MLLMs: PandaGPT, Emotion-LLaMA, and AffectGPT. %\textbf{Bold} indicates the best performance in each column. For all metrics, higher values indicate better performance.
``Mean'' represents the average performance across all datasets. We compare various mainstream MLLMs and apply our method to three baselines.
}
\label{tab:mer_bench}
\renewcommand{\arraystretch}{1} 
\setlength{\tabcolsep}{1.5pt} 
\resizebox{0.8\textwidth}{!}{
\begin{tabular}{lcccccccccc}
\toprule
\multirow{2}{*}{\textbf{Methods}} &
\multicolumn{4}{c}{\textbf{Basic}} &
\multicolumn{4}{c}{\textbf{Sentiment}} &
\multicolumn{1}{c}{\textbf{Fine-grained}} &
\multirow{2}{*}{\textbf{Mean}$\uparrow$} \\
\cmidrule(lr){2-5} \cmidrule(lr){6-9} \cmidrule(lr){10-10}
 & \textbf{MER23}$\uparrow$ & \textbf{MER24}$\uparrow$ & \textbf{MELD}$\uparrow$ & \textbf{IEMO}$\uparrow$
 & \textbf{MOSI}$\uparrow$ & \textbf{MOSEI}$\uparrow$ & \textbf{SIMS}$\uparrow$ & \textbf{SIMS v2}$\uparrow$
 & \textbf{OV-MERD+}$\uparrow$ &  \\

\hline
Video-LLaVA~\cite{lin2024video-llava} & 36.93 & 30.25 & 30.73 & 38.95 & 56.37 & 61.64 & 53.28 & 57.45 & 34.00 & 44.40 \\

Video-ChatGPT~\cite{maaz2024videochatgpt} & 44.86 & 46.80 & 37.33 & 56.83 & 54.42 & 63.12 & 64.82 & 65.80 & 39.80 & 52.64 \\

VideoChat2~\cite{li2024videochat2} & 33.67 & 54.50 & 36.64 & 48.70 & 66.84 & 54.32 & 69.49 & 70.66 & 39.21 & 52.67 \\

LLaMA-VID~\cite{li2024llama-vid} & 50.72 & 57.60 & 42.75 & 46.02 & 61.78 & 63.89 & 69.35 & 67.48 & 45.01 & 56.07 \\

VideoChat~\cite{li2023videochat} & 48.73 & 57.30 & 41.11 & 48.38 & 65.13 & 63.61 & 69.52 & 72.14 & 44.52 & 56.72 \\

Chat-UniVi~\cite{jin2024chatunivi} & 57.62 & 65.67 & 45.61 & 52.37 & 54.53 & 63.18 & 68.15 & 66.36 & 48.00 & 57.94 \\

mPLUG-Owl~\cite{ye2023mplugowl} & 56.86 & 59.89 & 49.11 & 55.54 & 72.40 & 72.91 & 72.13 & 75.00 & 48.18 & 62.45 \\

AVERE~\cite{chaubey2026avere} &56.87 &68.78&38.42&44.92&49.89&42.14&76.36&76.16&48.39&55.77\\

\hline
\rowcolor{white}
PandaGPT~\cite{su2023pandagpt} & 44.96 & 49.63 & 38.30 & 46.42 & 67.65 & 69.12 & 67.12 & 64.60 & 37.92 & 53.97  \\
\rowcolor{cyan!5}
\textbf{+Ours} & \textbf{48.35} & \textbf{50.61} & \textbf{39.54} & \textbf{49.33} & \textbf{68.70} & \textbf{69.83} & \textbf{69.35} & \textbf{66.35} & \textbf{39.05} & \textbf{55.68} \\

\rowcolor{white}
Emotion-LLaMA~\cite{cheng2024emotion-llama} & 64.91 & 66.14 & 63.46 & 47.30 & 63.60 & 71.83 & 67.67 & 72.88 & 46.80 & 62.73 \\
\rowcolor{cyan!5}
\textbf{+Ours} & \textbf{68.71} & \textbf{69.43} & \textbf{66.53} & \textbf{48.56} & \textbf{65.57} & \textbf{75.37} & \textbf{71.79} & \textbf{75.62} & \textbf{47.84} & \textbf{65.49} \\

\rowcolor{white}
AffectGPT~\cite{Lian2025AffectGPT} & 73.60 & 73.28 & 52.16 & 55.50 & 83.01 & 76.69 & 86.41 & 81.91 & 59.44 & 71.33 \\

\rowcolor{cyan!5}
\textbf{+Ours} & \textbf{76.35} & \textbf{73.62} & \textbf{53.78}  &  \textbf{56.17} & \textbf{85.13} & \textbf{77.05} &\textbf{88.45} & \textbf{85.54} & \textbf{61.04} & \textbf{73.01} \\
\bottomrule
\end{tabular}}
\end{table*}

\begin{table*}[t]
\centering
\small
\caption{Comparison on MME-Emotion \cite{zhang2025mme-emotion} for emotional understanding and reasoning. Results cover six emotional tasks and three evaluation metrics, all computed via a multi-agent system framework.} %. For all metrics, higher values indicate better performance.}
\label{tab:mme_emotion}
\renewcommand{\arraystretch}{1.2} 
\setlength{\tabcolsep}{1.5pt} 
\resizebox{0.8\textwidth}{!}{
\begin{tabular}{lccccccccccccccccccccc}
\toprule
\multirow{2}{*}{} & \multicolumn{3}{c}{\textbf{ER-Lab}} & \multicolumn{3}{c}{\textbf{FG-SA}} & \multicolumn{3}{c}{\textbf{IR}} & \multicolumn{3}{c}{\textbf{ML-ER}} & \multicolumn{3}{c}{\textbf{Noise-ER}} & \multicolumn{3}{c}{\textbf{SA}} & \multicolumn{3}{c}{\textbf{Mean}} \\
\cmidrule(lr){2-4} \cmidrule(lr){5-7} \cmidrule(lr){8-10} \cmidrule(lr){11-13} \cmidrule(lr){14-16} \cmidrule(lr){17-19} \cmidrule(lr){20-22}
\textbf{Methods} & Rec & Rea & CoT & Rec & Rea & CoT & Rec & Rea & CoT & Rec & Rea & CoT & Rec & Rea & CoT & Rec & Rea & CoT & Rec & Rea & CoT \\
\hline
PandaGPT~\cite{su2023pandagpt} 
& 15.5 & 42.6 & 30.6 & 34.8 & 47.8 & 41.3 & 13.8 & 48.5 & 31.2 & 2.0 & 37.8 & 21.4 & 21.2 & 40.9 & 31.3 & 35.2 & 51.0 & 43.1 & 20.4 & 44.8 & 33.2 \\
\rowcolor{cyan!5}
\textbf{+Ours} 
& \textbf{19.7} & \textbf{45.6} & \textbf{31.1} 
& \textbf{37.0} & \textbf{49.4} & \textbf{43.2} 
& \textbf{15.8} & \textbf{48.8} & \textbf{32.3} 
& \textbf{5.0} & \textbf{42.0} & \textbf{22.0} 
& \textbf{22.0} & \textbf{41.5} & \textbf{31.4} 
& \textbf{41.2} & \textbf{54.5} & \textbf{47.9} 
& \textbf{23.5} & \textbf{47.0} & \textbf{34.7} \\
\hline
Emotion-LLaMA~\cite{cheng2024emotion-llama} 
& 11.7 & 32.8 & 22.3 & 26.8 & 54.4 & 41.0 & 11.0 & 45.1 & 28.0 & 8.3 & 46.9 & 27.6 & 44.0 & 60.3 & 52.4 & 36.2 & 61.5 & 48.9 & 23.0 & 50.2 & 36.7 \\
\rowcolor{cyan!5}
\textbf{+Ours} 
& \textbf{14.8} & \textbf{37.5} & \textbf{26.2} 
& \textbf{29.0} & \textbf{55.2} & \textbf{41.7} 
& \textbf{13.8} & \textbf{47.3} & \textbf{30.6} 
& \textbf{8.7} & \textbf{49.9} & \textbf{29.3} 
& \textbf{45.6} & \textbf{60.8} & \textbf{52.9} 
& \textbf{36.5} & \textbf{63.6} & \textbf{50.1} 
& \textbf{24.7} & \textbf{52.4} & \textbf{38.5} \\
\hline
AffectGPT~\cite{Lian2025AffectGPT} 
& 16.2 & 54.2 & 35.2 & 15.6 & 63.5 & 39.6 & 3.0 & 50.4 & 26.7 & 0.6 & 49.9 & 25.6 & 28.8 & 65.3 & 47.0 & 25.5 & 67.6 & 46.7 & 15.0 & 58.5 & 36.8 \\
\rowcolor{cyan!5}
\textbf{+Ours} 
& \textbf{19.3} & \textbf{59.6} & \textbf{39.5} & \textbf{18.6} & \textbf{68.4} & \textbf{43.5} & \textbf{5.2} & \textbf{54.5} & \textbf{29.9} & \textbf{0.7} & \textbf{54.6} & \textbf{27.6} & \textbf{31.8} & \textbf{67.4} & \textbf{49.6} & 25.5 & \textbf{70.7} & \textbf{48.1} & \textbf{16.9} & \textbf{62.5} & \textbf{39.7} \\
\bottomrule
\end{tabular}
}
\end{table*}

\section{Experiments}
\label{sec:experiments}

\subsection{Experimental Setting}
\label{sec:settings}
\subsubsection{\textit{\textbf{Datasets and Metrics}}}
We evaluate our method on two multimodal emotion benchmarks: MER-UniBench~\cite{Lian2025AffectGPT} and MME-Emotion~\cite{zhang2025mme-emotion}.
\textbf{MER-UniBench} is a comprehensive evaluation benchmark designed to cover typical multimodal emotion recognition tasks, including basic emotion recognition, sentiment analysis, and fine-grained emotion recognition. For the three tasks, we select nine datasets: MER2023~\cite{lian2023mer}, MER2024~\cite{lian2024mer}, MELD~\cite{poria2019meld}, IEMOCAP~\cite{busso2008iemocap}, CMU-MOSI~\cite{zadeh2016SMUMOSI}, CMU-MOSEI~\cite{bagher-zadeh-etal-2018-multimodal}, CH-SIMS~\cite{yu2020chCh-sims}, CH-SIMS v2~\cite{liu2022makeCh-sims-v2}, and OV-MERD+~\cite{lian2025ovmer}. We adopt the official evaluation metrics provided by MER-UniBench, tailored for the free-form natural language output style of MLLMs.
% To assess the emotion recognition capability of our method, we comprehensively conduct experiments on seven widely used multimodal emotion recognition datasets: 
% Following the experimental settings in~\cite{Lian2025AffectGPT}, we adopt specific evaluation metrics for each dataset. Specifically, \textit{hit-rate} for MER2023, MER2024, MELD, and IEMOCAP; \textit{weighted average F1-score (WAF)} for CH-SIMS and CH-SIMS v2; and the \textit{emotion-wheel} metric for OV-MERD. 
\textbf{MME-Emotion} is the largest emotional intelligence benchmark for MLLMs, which assesses both emotional understanding and reasoning capabilities. It contains 6,500 video clips with task-specific question-answering (QA) pairs, spanning 27 distinct scenario types. In this work, we report results on six emotional tasks, including emotion recognition in the lab (ER-Lab), %emotion recognition in the wild (ER-Wild), 
emotion recognition under noise (Noise-ER), multi-label emotion recognition (ML-ER), sentiment analysis (SA), fine-grained sentiment analysis (FG-SA), and intent recognition (IR). %The visual clues, extracted audio clues, groundtruth emotion labels, and partitioned answer steps of a specific MLLM are fed into a GPT-based judge agent to evaluate the performance using three metrics: recognition score (Rec-S), reasoning score (Rea-S), and Chain-of-Thought score (CoT-S).
Evaluation metrics include recognition (Rec) score, reasoning (Rea) score, and Chain-of-Thought (CoT) score, which are computed via a multi-agent system framework.

\subsubsection{\textit{\textbf{Implementation Details}}}
% The complete pipeline of \texttt{LAMR} is summarized in Algorithm~\ref{alg:LAMP}.
For layer profiling, we construct a small validation set $\mathcal{D}_{val}$ by combining the CMU-MOSEI and CMU-MOSI datasets~\cite{bagher-zadeh-etal-2018-multimodal}, which are widely used benchmarks for multimodal sentiment analysis. This validation set is used to estimate layer-wise attention statistics and determine the layer threshold $\mathcal{S}_{\ell^*}$.
We apply our method to three representative MER baselines: PandaGPT~\cite{su2023pandagpt}, Emotion-LLaMA~\cite{cheng2024emotion-llama}, and AffectGPT~\cite{Lian2025AffectGPT}. The proposed attention rebalance strategy is performed only during inference on the test set $\mathcal{D}_{test}$.
For the two types of layers identified by our method, the rebalance hyperparameters $(\alpha_s, \alpha_d)$ are set to $(0.1, 0.1)$ for AffectGPT, and $(0.2, 0.2)$ for the other baselines, respectively. For LLM-based evaluation on MME-Emotion, we employ \texttt{gpt-4.1-2025-04-14} and \texttt{gpt-4o-2024-11-20} as judge models. 
% \textbf{Please see the Appendix~\ref{appen:implementation details} for more implementation details.}

%%%%%%%%%%%%%%%%%%%%%%%%%%%%%%%%%%%%%%%%%%%%%%%%%%%%%%%%%%%%%%%%%
%%%%%%%%%%%%%%%%%%%%%%%%%%%  主实验  %%%%%%%%%%%%%%%%%%%%%%%%%%%%%
\subsection{Performance Comparison}
\label{main results}

\subsubsection{\textit{\textbf{Emotion Recognition}}}
For a comprehensive comparison, we include representative generalist MLLMs (\eg, Video-LLaVA~\cite{lin2024video-llava}) and specialized MER models (\eg, AVERE~\cite{chaubey2026avere}), and apply our method to three baseline architectures. As shown in Table~\ref{tab:mer_bench}, applying \texttt{LAMR} to PandaGPT, Emotion-LLaMA, and AffectGPT results in consistent performance improvements across all nine datasets, with average gains of $1.71\%$, $2.76\%$, and $1.68\%$, respectively.
This indicates that our inference-time rebalancing strategy is robust and model-agnostic. Notably, these gains are observed not only on basic emotion recognition and sentiment analysis, but also on fine-grained emotion recognition tasks (OV-MERD+ dataset), which require more delicate cross-modal evidence aggregation. \texttt{LAMR} improves the multimodal reasoning process by alleviating unimodal dominance and strengthening cross-modal integration. 

\begin{table}[t]
\centering
\footnotesize
\caption{Ablation study on layer partition strategies. %We analyze the impact of selecting different layers as integration layers. 
``Shallow'', ``Middle'', and ``Deep'' denote static heuristic partitions that evenly divide the layers into three contiguous stages. We designate one or two of these stages as integration layers. %``Ours'' utilizes our Adaptive Layer Role Profiling strategy.
}
\label{tab:ablation_layer}
\setlength{\tabcolsep}{6pt}
\begin{tabular}{lcccc}
\toprule
\textbf{Strategy} & \textbf{MELD} & \textbf{IEMOCAP} & \textbf{SIMS} & \textbf{Mean} \\
\hline
AffectGPT  & 52.16 & 55.50 & 86.41 & 64.69 \\
\hline
\rowcolor{gray!10}\multicolumn{5}{l}{\textit{Random Assignment}} \\
30\% Integration & \textbf{53.99} & 51.53 & 80.44 & 61.99 \\
50\% Integration & 49.71 & 47.51 & 80.79 & 59.34 \\
70\% Integration & 47.15 & 46.28 & 80.68 & 58.04 \\
\hline
\rowcolor{gray!10}\multicolumn{5}{l}{\textit{Static Assignment}} \\
Shallow & 51.45 & 51.51 & 82.88 & 61.95 \\
Middle & 53.52 & 52.44 & 83.15 & 63.04 \\
Deep & 52.96 & 51.86 & 81.04 & 61.95 \\
Shallow + Middle & 52.56 & 51.53 & 83.77 & 62.62 \\
Middle + Deep & 52.70 & 52.23 & 80.60 & 61.84 \\
Shallow + Deep & 51.29 & 50.95 & 82.16 & 61.47 \\
\hline
\rowcolor{cyan!5} \textbf{Ours} & 53.78 & \textbf{56.17} & \textbf{88.45} & \textbf{66.13} \\
\bottomrule
\end{tabular}
\vspace{-1em}
\end{table}

\subsubsection{\textit{\textbf{Emotional Intelligence}}}
Beyond recognition accuracy, we evaluate \texttt{LAMR} on the MME-Emotion benchmark for holistic emotional intelligence. As reported in Table~\ref{tab:mme_emotion}, \texttt{LAMR} consistently improves performance across six emotional tasks, with clear gains on reasoning-related metrics. For example, on the FG-SA task, AffectGPT's Rec score increases from 15.6 to 18.6, while CoT score rises from 39.6 to 43.5. \texttt{LAMR} enhances not only the final predictions but also the coherence and logical structure of the reasoning process. %These improvements are attributed to \texttt{LAMR}’s ability to operate at the level of internal multimodal evidence aggregation in emotion reasoning.
These improvements are attributed to \texttt{LAMR} operating at the level of internal multimodal evidence aggregation, rather than merely adjusting output-level predictions.

% These findings imply that by rectifying attention imbalance, \texttt{LAMR} effectively ensures that the emotional reasoning is supported by a balanced integration of multimodal evidence, leading to more robust emotional intelligence.

\begin{table}[!t]
\centering
\footnotesize
\caption{Ablation study on the quantitative metrics used for adaptive layer role profiling, \ie, CMI and ME.}
\label{tab:ablation_metrics}
\begin{tabular}{cc|cccc}
\toprule
\textbf{CMI} & \textbf{ME} & \textbf{MELD} & \textbf{IEMOCAP} & \textbf{SIMS} & \textbf{Mean} \\
\hline
\ding{51} & \ding{55} & 52.25 & 53.62 & 83.67 & 63.18 \\
\ding{55} & \ding{51} & 51.60 & 53.18 & 85.47 & 63.42 \\
\rowcolor{cyan!5} \ding{51} & \ding{51} & \textbf{53.78} & \textbf{56.17} & \textbf{88.45} & \textbf{66.13} \\
\bottomrule
\end{tabular}
\vspace{-1em}
\end{table}

\begin{table}[t]
\centering
\footnotesize
\caption{Ablation study on layer-wise attention rebalance. ``Reversed Intervention'' denotes exchanging intervention strategies for two types of layers.}
\label{tab:ablation_rebalance}
\setlength{\tabcolsep}{6pt}
\begin{tabular}{lcccc}
\toprule
\textbf{Strategy} & \textbf{MELD} & \textbf{IEMOCAP} & \textbf{SIMS} & \textbf{Mean} \\
\hline
AffectGPT  & 52.16 & 55.50 & 86.41 & 64.69 \\
\hline
Uniform Propagation & 47.75 & 51.23 & 81.53 & 60.17 \\
Uniform Equilibrium & 53.53 & 51.11 & 83.92 & 62.85 \\
Reversed Intervention & 53.28 & 51.44 & 84.26 & 62.99\\
\hline
\rowcolor{cyan!5} \textbf{Ours} & \textbf{53.78} & \textbf{56.17} & \textbf{88.45} & \textbf{66.13} \\
\bottomrule
\end{tabular}
\vspace{-1em}
\end{table}

% \begin{figure}[!t]
%     \centering
%     \includegraphics[width=0.48\textwidth]{Figs/example-fig1.pdf}
%         \caption{Revisiting the motivating example from Figure~\ref{fig:introduction}. }
%     %Our method facilitates balanced cross-modal interaction, enabling the model to correct initial misjudgments and generate accurate affective reasoning.}
%     \label{fig:example-fig}
% \end{figure}

\subsection{Ablation Studies}
\label{sec:ablation}

\subsubsection{\textit{\textbf{Impact of Adaptive Layer Role Profiling}}}
To validate the necessity of adaptive layer role profiling, we compare \texttt{LAMR} with two alternative strategies: \textit{Random Assignment}, which randomly selects 30\%, 50\%, or 70\% of the layers as integration layers, and \textit{Static Assignment}, which heuristically partitions layers into shallow, middle, and deep stages and applies integration to one or two fixed stages. As shown in Table~\ref{tab:ablation_layer}, both strategies reduce the baseline performance. In particular, random assignment leads to the most severe performance degradation, as it blindly applies inappropriate interventions, disrupting the internal organization of multimodal processing.
In contrast, our adaptive profiling strategy yields the best results, demonstrating that accurate identification of layer-specific functions is critical for effective multimodal attention rebalancing.

% indicating that layer functional roles cannot be captured by fixed heuristics. Conversely, our strategy achieves a superior mean score of $66.13$, demonstrating the necessity for layer characterization in multimodal emotion reasoning.

\subsubsection{\textit{\textbf{Impact of CMI and ME}}}
Our layer role profiling relies on a unified layer score $\mathcal{S}_\ell$ that jointly combines CMI and ME. To assess the contribution of each metric, we compare \texttt{LAMR} with two ablated variants that rely on a single metric. As shown in Table~\ref{tab:ablation_metrics}, both single-metric strategies lead to suboptimal performance, with the mean score dropping from $66.13$ to $63.18$ (only CMI) and $63.42$ (only ME), respectively. These results validate our design motivation that only layers exhibiting both high cross-modal intensity and high modality entropy should be identified as integration layers for cross-modal semantic aggregation.
%Specifically, CMI fails to capture modality bias, whereas ME ignores interaction intensity. Therefore, combining both is essential to pinpoint layers that feature simultaneous cross-modal integration and balanced participation.

\subsubsection{\textit{\textbf{Impact of Layer-wise Attention Rebalance}}} 
To examine whether the effectiveness of \texttt{LAMR} stems from matching intervention types to layer-specific functions, we test three variants: {Uniform Propagation}, which applies diffusive propagation to all layers; {Uniform Equilibrium}, which applies modality equilibrium to all layers; and {Reversed Intervention}, which swaps the two strategies between stabilization and integration layers. As shown in Table~\ref{tab:ablation_rebalance}, all three variants lead to significant performance degradation. These results demonstrate that effective attention rebalancing requires alignment with the intrinsic functional role of each layer, thereby confirming the necessity of our layer-wise intervention design.

% We evaluate layer-wise attention rebalance against three strategies: \textit{Uniform Propagation}, \textit{Uniform Equilibrium}, and \textit{Reversed Intervention}. As shown in Table~\ref{tab:ablation_rebalance}, the consistent performance degradation caused by disregarding layer heterogeneity underscores the imperative of aligning interventions with intrinsic layer roles.

\begin{table}[t]
\footnotesize
\centering
\caption{Performance comparison on the CA-MER benchmark~\cite{han2025benchmarking} under realistic emotion conflicts. The three subsets---video-aligned, audio-aligned, and consistent---represent settings in which the video modality, the audio modality, or all modalities reflect the true emotion, respectively. %, which comprises three subsets: video-aligned, audio-aligned, and consistent. The video- and audio-aligned subsets comprise samples where the respective modality reflects the true emotion, while the others present conflicting cues.
}
\label{tab:camer}

\setlength\tabcolsep{3pt}
\begin{tabular}{l ccc ccc ccc}
\toprule
\multirow{2}{*}{\textbf{Model}} 
 & \multicolumn{3}{c}{\textbf{Video-aligned}} 
 & \multicolumn{3}{c}{\textbf{Audio-aligned}} 
 & \multicolumn{3}{c}{\textbf{Consistent}} 
\\ 
\cmidrule(lr){2-4}\cmidrule(lr){5-7}\cmidrule(lr){8-10}
 & Acc & Recall & Avg 
 & Acc & Recall & Avg 
 & Acc & Recall & Avg 
\\ 
\hline

AffectGPT 
&44.92 &50.80 &47.86
&52.26 &55.54 &53.90
&51.86 &56.90 &54.38
% & 0.4968 & 0.5441 & 0.5210 
\\

\rowcolor{cyan!5}
\textbf{+ Ours} 
&\textbf{45.89} & \textbf{52.66} & \textbf{49.28}
&\textbf{52.56} & \textbf{60.00} & \textbf{56.28}
&\textbf{52.72} & \textbf{59.84} & \textbf{56.28}
% & \textbf{0.5039} & \textbf{0.5750} & \textbf{0.5390} 
\\
\bottomrule
\end{tabular}
\vspace{-1em}
\end{table}

\begin{table}[t]
\centering
\footnotesize
\caption{Comparison with other training-free rebalancing methods on MER-UniBench.}
\label{tab:affectgpt_comparison}
\setlength{\tabcolsep}{3pt}
\begin{tabular}{lccccccc}
\toprule
\textbf{Methods} &
\textbf{MER23} & \textbf{MER24} & \textbf{MELD} & \textbf{IEMO} & \textbf{SIMS} & \textbf{SIMSv2} &
\textbf{OV.}  \\
\hline
AffectGPT
& 73.60 & 73.28 & 52.16 & 55.50 & 86.41 & 81.91 & 59.44  \\
+ Devils~\cite{jiang2025devils}
& 75.27 & 70.13 & 52.34 & 49.40 & 79.93 & 81.95 & 59.80 \\
+ AVCD~\cite{jung2025avcd}
& \underline{75.35} & \underline{75.05} & \underline{53.56} & 51.83 & 80.97 & 82.43 & \underline{60.69}  \\
+ LISA~\cite{guo2025lisa}
& 73.14 & \textbf{75.79} & 52.88 & \underline{55.53} & \underline{86.43} & \underline{85.41} & 60.33  \\
\rowcolor{cyan!5}
\textbf{+ Ours} 
& \textbf{76.35} & 73.62 & \textbf{53.78} & \textbf{56.17} & \textbf{88.45} & \textbf{85.54} & \textbf{61.04}  \\
\bottomrule
\end{tabular}
\vspace{-1em}
\end{table}

\subsection{Further Analysis}

\subsubsection{\textit{\textbf{Robustness Under Emotion Conflicts}}}
CA-MER~\cite{han2025benchmarking} is a benchmark designed to examine MLLMs under realistic emotion conflicts. It constitutes a challenging setting for MER, where only one or all modalities reflect the true emotion.  
As shown in Table~\ref{tab:camer}, AffectGPT performs worse on the Video-aligned subset compared to the Audio-aligned subset, indicating a clear bias toward dominant modalities. By re-balancing attention at inference time, \texttt{LAMR} effectively mitigates modality dominance %under conflict 
and improves performance across all subsets, achieving a $1.9\%$ gain in average score.

% we observe that the Video-aligned subset achieves lower accuracy than the Audio-aligned and Consistent subsets. This suggests that the model struggles to effectively incorporate information from the non-dominant modality, consistent with the observations in Figure~\ref{fig:introduction}.

% Devils in Middle Layers of Large Vision-Language Models: Interpreting, Detecting and Mitigating Object Hallucinations via Attention Lens~\cite{jiang2025devils} 
% LISA: A Layer-wise Integration and Suppression Approach for Hallucination Mitigation in Multimodal Large Language Models~\cite{guo2025lisa}
% AVCD: Mitigating Hallucinations in Audio-Visual Large Language Models through Contrastive Decoding~\cite{jung2025avcd} 

\subsubsection{\textit{\textbf{Comparison with Training-Free Rebalancing Methods}}}
To evaluate the efficacy of existing training-free rebalancing approaches for MER, we re-implement three representative methods, Devils~\cite{jiang2025devils}, AVCD~\cite{jung2025avcd}, and LISA~\cite{guo2025lisa}, on the same backbone following their original intervention principles for fair comparison. These methods encompass diverse rebalancing paradigms: AVCD performs contrastive decoding, Devils focuses on modality token-level attention enhancement, and LISA implements a layer-wise integration and suppression approach based on predefined layer functions.
As shown in Table~\ref{tab:affectgpt_comparison}, these methods often yield unstable performance when transferred to MER: while moderate gains are observed on some datasets, performance degrades on others. This suggests that generic interventions, which rely on fixed layer priors or non-adaptive modality adjustments, fail to account for the fine-grained functional roles of layers in multimodal emotion reasoning. In contrast, \texttt{LAMR} achieves consistent improvements by explicitly grounding its observation and intervention in MER.

% We benchmark against state-of-the-art training-free strategies designed for modality rebalance: Devils~\cite{jiang2025devils}, AVCD~\cite{jung2025avcd}, and LISA~\cite{guo2025lisa}, which primarily intervene based on fixed layer-wise topological priors or indiscriminate masking non-dominant modalities. 
%Specifically, Devils and LISA intervene based on fixed topological priors, while AVCD applies indiscriminate masking to non-dominant modalities to offset prior bias. 
% We implement these methods on the strongest baseline, AffectGPT, for a fair comparison. As detailed in Table~\ref{tab:affectgpt_comparison}, these intervention methods exhibit instability. This suggests that such coarse-grained regulation risks discarding subtle emotional cues, whereas our method succeeds by aligning interventions with the MER model's intrinsic functional heterogeneity instead of imposing rigid priors.

\begin{table}[!t]
\centering
\footnotesize
\caption{Scalability analysis across different model sizes ($1.5$B, $7$B, and $14$B).}
\label{tab:results_comparison}
\begin{tabular}{lcccc}
\toprule
\textbf{Model Setup} & \textbf{MER24}& \textbf{MELD} & \textbf{IEMOCAP}  & \textbf{Avg.\ $\Delta$} \\
\hline

% \rowcolor{gray!10}
% \textit{1.5B Parameters} & & & & \\
AffectGPT-1.5B &75.39  &49.34 & 52.32 & -- \\
\rowcolor{cyan!5}
\textbf{+ Ours} & 76.68 & 49.97 & 52.84 & {\color{red}0.81$\uparrow$} \\

AffectGPT-7B & 73.28 & 52.16 & 55.50  & -- \\
\rowcolor{cyan!5}
\textbf{+ Ours} &73.62& 53.78 & 56.17 & {\color{red}0.88$\uparrow$} \\

% \rowcolor{gray!10}
% \textit{14B Parameters} & & & & \\
AffectGPT-14B &77.36 & 55.01 & 60.80 & -- \\
\rowcolor{cyan!5}
\textbf{+ Ours} &80.02 & 55.57 & 65.44 & {\color{red}2.62$\uparrow$} \\
\bottomrule
\end{tabular}
% \vspace{-1em}
\end{table}

\begin{figure}[t!] % h=here, t=top, b=bottom, p=page
    \centering
    \includegraphics[width=0.5\textwidth]{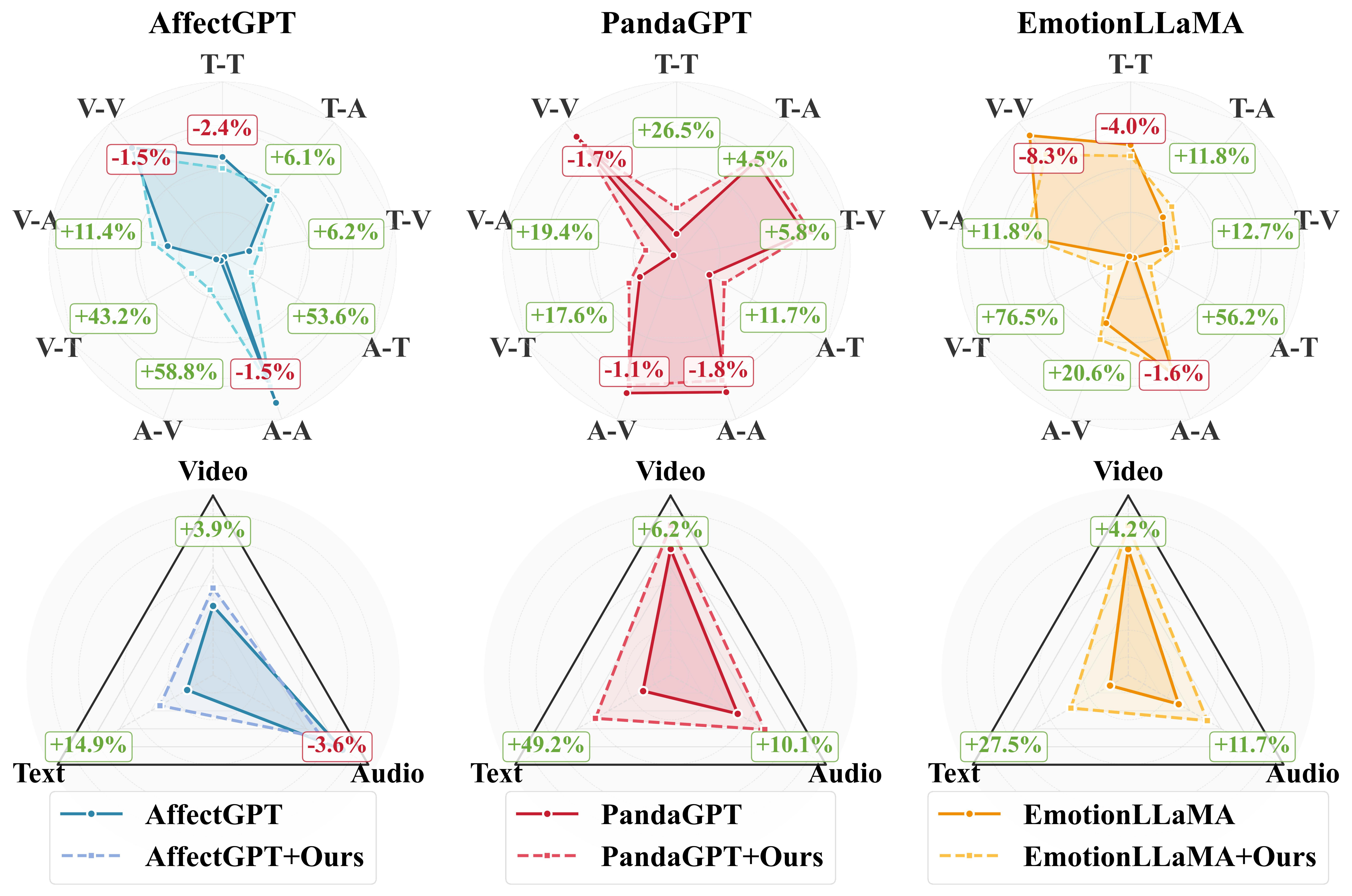}
    \caption{Effect of \texttt{LAMR} on modality-level and cross-modal attention rebalancing. (a/b) Modality-wise and cross-modal attention allocation with \texttt{LAMR}. Percentages indicate changes relative to the baseline.
%\textbf{Top:} Comparison of mean modality attention distributions. \textbf{Bottom:} Comparison of mean cross attention distributions. Percentages indicate the relative changes in attention from the original baseline to our method.
    }
    \Description{Radar charts compare modality-wise and cross-modal attention before and after LAMR for AffectGPT, PandaGPT, and EmotionLLaMA, with percentage changes relative to each baseline.}
    \label{fig:radar}
    \vspace{-1em}
\end{figure}

\subsubsection{\textit{\textbf{Scalability Across Model Sizes}}} 
In Table~\ref{tab:results_comparison}, we evaluate \texttt{LAMR} on AffectGPT backbones of increasing capacity ($1.5$B, $7$B, and $14$B parameters). \texttt{LAMR} consistently yields performance gains across all model scales, indicating that modality imbalance persists even in large-capacity MLLMs and cannot be resolved solely by scaling up model size. More importantly, \texttt{LAMR} is scalable and benefits both lightweight and large-scale backbones. 
% Our method consistently improves performance across all backbone sizes, demonstrating that the proposed inference-time strategy is robust to model capacity and effective for both lightweight and large-scale MLLMs.

\subsubsection{\textit{\textbf{Results of Attention Rebalance}}} 
Figure~\ref{fig:radar} provides a quantitative view of how \texttt{LAMR} corrects both facets of modality imbalance. %Figure~\ref{fig:radar} quantitatively validates the effectiveness of our method in mitigating the modality imbalance.
As shown in Figure~\ref{fig:radar} (a), \texttt{LAMR} significantly balances the attention allocated to three modalities. In AffectGPT, attention to the non-dominant text and video modalities increases by $+14.9\%$ and $+3.9\%$, respectively, while attention to the dominant audio modality decreases by $-3.6\%$. Figure~\ref{fig:radar} (b) shows an increase in cross-modal attention relative to self-modal attention. In particular, AffectGPT exhibits a $+58.8\%$ gain in audio-to-video attention, and EmotionLLaMA shows a $+76.5\%$ increase in video-to-text interaction, indicating that \texttt{LAMR} actively promotes inter-modal information propagation. Overall, \texttt{LAMR} corrects both unimodal over-reliance and cross-modal under-utilization, directly addressing the two-sided attention dilemma identified in our analysis. 
%Regarding unimodal over-reliance, Figure~\ref{fig:radar} (a) demonstrates the weights for non-dominant modalities are elevated, evidenced by a relative increase of $49.2\%$ and $27.5\%$ in Text attention for PandaGPT and EmotionLLaMA, respectively. For cross-modal under-utilization, Figure~\ref{fig:radar} (b) reveals an increase in the attention weights of cross-modal pairs relative to self-modal ones, with AffectGPT showing a $58.8\%$ gain in Audio-to-Video attention and EmotionLLaMA exhibiting a $76.5\%$ surge in Video-to-Text interaction.  
%This global elevation of inter-modal values indicates that the model effectively promotes information interaction across distinct modalities.

% \noindent \textbf{Case study.}
% To provide intuitive insights into our method’s efficacy, Figure~\ref{fig:example-fig} revisits the motivating example introduced in Section~\ref{sec:intro}. While the baseline model exhibits uni-modal over-reliance and cross-modal under-utilization, leading to misaligned affective reasoning, our approach effectively mitigates this issue. Guided by the layer hierarchy identified by Adaptive Layer Role Profiling, LAR stabilizes attention via Modality Anchoring Equilibrium and enhances cross-modal information flow through Intermodal Diffusive Propagation. This layer-wise refinement enables the model to better ground reasoning in multimodal evidence and capture subtle emotional cues that were previously underutilized. More cases can be found at Appendix~\ref{appen:qualitative_results}

\subsection{Qualitative Results}
\begin{figure}[t!] % h=here, t=top, b=bottom, p=page
    \centering
    \includegraphics[width=0.48\textwidth]{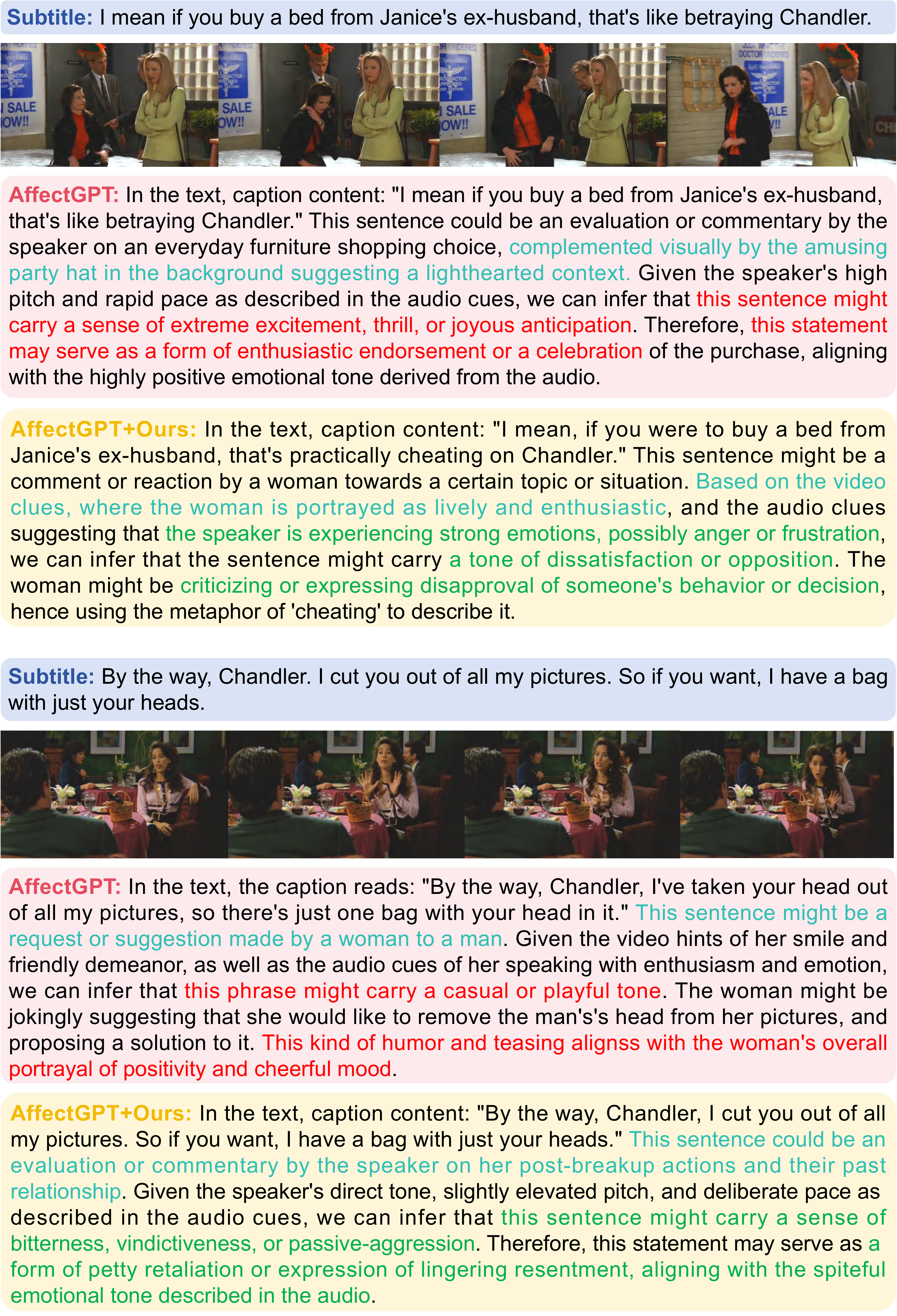}
    \caption{Two challenging examples of modality imbalance. Cyan, green, and red fonts indicate shared correct, correct, and erroneous interpretations, respectively. %shared correct interpretations by both the baseline and our method, the accurate interpretations, and the erroneous interpretations
    }
    \Description{Two dialogue examples compare AffectGPT with AffectGPT plus LAMR. Colored text marks interpretations that are shared and correct, newly correct, or erroneous, showing how LAMR reduces modality-driven reasoning errors.}
    \label{fig:casestudy}
    \vspace{-2em}
\end{figure}

As shown in Figure~\ref{fig:casestudy}, we examine two challenging cases that reflect distinct forms of modality imbalance in multimodal emotion reasoning. 
%LAMR effectively mitigates two distinct modality imbalance issues. 
In the first case of unimodal over-reliance, the baseline overemphasizes acoustic cues (\eg, \textit{high pitch}) and erroneously predicts \textit{excitement}. %erroneously predicts excitement by heavily favoring audio cues like high pitch, marginalizing the actual text. Our method suppresses this dominance to correctly infer dissatisfaction. 
In the second case of cross-modal under-utilization, the baseline fails to reconcile positive visual/audio signals (\ie, visual smile/enthusiastic tone) with hostile text, resulting in a superficial interpretation of \textit{playful}. 
%the baseline interprets a smile and enthusiastic tone as playful, failing to integrate the hostile textual semantics. 
In contrast, LAMR achieves more nuanced emotional understanding by suppressing excessive modality reliance and enhancing cross-modal interaction.
%By actively synthesizing these isolated cues, LAMR accurately detects underlying passive-aggression, demonstrating significantly deeper emotional understanding. 
% Additional qualitative results are provided in Appendix~\ref{appen:qualitative_results}.

%%%%%%%%%%%%%%%%%%%%%%%%%%%%%%%%%%%%%%%%%%%%%%%%%%%%%%%%%%%%%%
\section{Conclusion}
In this work, we aim to address the modality imbalance problem in multimodal emotion reasoning, manifested as unimodal over-reliance and cross-modal under-utilization. We propose Layer-Aware Modality Rebalancing (\texttt{LAMR}), which rebalances multimodal attention through adaptive layer role profiling and role-conditioned attention reallocation during inference. Extensive experiments on multiple MER benchmarks demonstrate that \texttt{LAMR} effectively mitigates modality imbalance and consistently improves emotion reasoning performance across diverse MLLM backbones without requiring any additional training.

\begin{acks}
This work was supported by the National Natural Science Foundation of China (62402471, U22A2094), the Yangtze River Delta Science and Technology Innovation Community Joint Research (Basic Research) Project (2025CSJZN01600), and the National Science and Technology Major Project (2025ZD0215301).
\end{acks}

\clearpage
\bibliographystyle{ACM-Reference-Format}
\balance
\bibliography{example_paper}

%%%%%%%%%%%%%%%%%%%%%%%%%%%%%%%%%%%%%%%%%%%%%%%%%%%%%%%%%%%%%%%%%%%%%%%%%%%%%%%
%%%%%%%%%%%%%%%%%%%%%%%%%%%%%%%%%%%%%%%%%%%%%%%%%%%%%%%%%%%%%%%%%%%%%%%%%%%%%%%
% APPENDIX
%%%%%%%%%%%%%%%%%%%%%%%%%%%%%%%%%%%%%%%%%%%%%%%%%%%%%%%%%%%%%%%%%%%%%%%%%%%%%%%
%%%%%%%%%%%%%%%%%%%%%%%%%%%%%%%%%%%%%%%%%%%%%%%%%%%%%%%%%%%%%%%%%%%%%%%%%%%%%%%

\end{document}